\documentclass[letterpaper, 10 pt, conference]{ieeeconf}  

\IEEEoverridecommandlockouts                              

\usepackage{graphics} 
\usepackage{epsfig} 
\usepackage{mathptmx} 
\usepackage{times} 
\usepackage{amsmath} 
\usepackage{amssymb}  
\usepackage{stmaryrd} 
\usepackage{multirow} 
\usepackage{graphicx}
\usepackage{subcaption} 
\usepackage{wrapfig}
\usepackage{amsmath}
\usepackage{bbm}
\usepackage{booktabs}
\usepackage{verbatim}

\newcommand\figref{Fig.~\ref}
\newcommand\tabref{Tab.~\ref}

\renewcommand{\baselinestretch}{0.99}

\title{\LARGE \bf
Learning Interactive Driving Policies via Data-driven Simulation
}

\author{Tsun-Hsuan Wang$^{1, *}$, Alexander Amini$^{1, *}$, Wilko Schwarting$^{1}$, Igor Gilitschenski$^{2}$,\\Sertac Karaman$^{1}$, and Daniela Rus$^{1}$
\thanks{* The first two authors have contributed equally to this work. This work was supported by National Science Foundation and Toyota Research Institute. We gratefully acknowledge the support of NVIDIA with the donation of the Drive AGX Pegasus.}
\thanks{$^{1}$ {\tt\{tsunw,amini,wilkos,sertac,rus\}@mit.edu} Department of Electrical Engineering and Computer Science, Massachusetts Institute of Technology (MIT).}%
\thanks{$^{2}$ {\tt gilitschenski@cs.utoronto.edu} Department of Computer Science, University of Toronto and Toyota Research Institute (TRI).}%
}

\begin{document}

\maketitle
\thispagestyle{empty}
\pagestyle{empty}

\begin{abstract}
Data-driven simulators promise high data-efficiency for driving policy learning.
When used for modelling interactions, this data-efficiency becomes a bottleneck: Small underlying datasets often lack interesting and challenging edge cases for learning interactive driving. We address this challenge by proposing a simulation method that uses in-painted ado vehicles for learning robust driving policies. Thus, our approach can be used to learn policies that involve multi-agent interactions and allows for training via state-of-the-art policy learning methods. We evaluate the approach for learning standard interaction scenarios in driving. In extensive experiments, our work demonstrates that the resulting policies can be directly transferred to a full-scale autonomous vehicle without making use of any traditional sim-to-real transfer techniques such as domain randomization.

\end{abstract}

\section{Introduction}
\label{sec:introduction}


Intelligent agents can achieve complex continuous control and decision making in the presence of rich multi-agent interactions as well as diverse lighting and environmental conditions. This ability requires learning representations from raw perception to high-level control actions. The interactive multi-agent case is challenging for autonomous navigation. End-to-end policy learning has demonstrated great promise for lane-stable single-agent navigation. However, to date, these networks are limited to simplistic road environments~\cite{Codevilla2018, Kendall2018}, navigation with no interactions~\cite{Bojarski2016, Amini2019}, testing in solely passive settings~\cite{Xu2017, Hubschneider2019}, or are deployed only in simulated environments disregarding real-world transferability~\cite{Rhinehart2018, Mehta2018, Xiao2020}. 

End-to-end learning of multi-agent visual control policies will increase the abilities of autonomous agents to reason and make decisions about how to move in interactive environments. End-to-end imitation learning requires capturing expert training data from extensive edge cases, such as recovery from off-orientation positions or near collisions with other agents. This is prohibitively expensive and dangerous for interactive situations. Simulation presents a solution to efficiently training and testing autonomous agents before deploying them in the real-world. Sim-to-real transfer has achieved great results in constrained problem settings~\cite{Pan2017, Bewley2019, Amini2020}, but it remains challenging for interactive autonomous driving and other interactive robotics tasks. 

\begin{figure}[!t]
\centering
\includegraphics[width=0.8\linewidth]{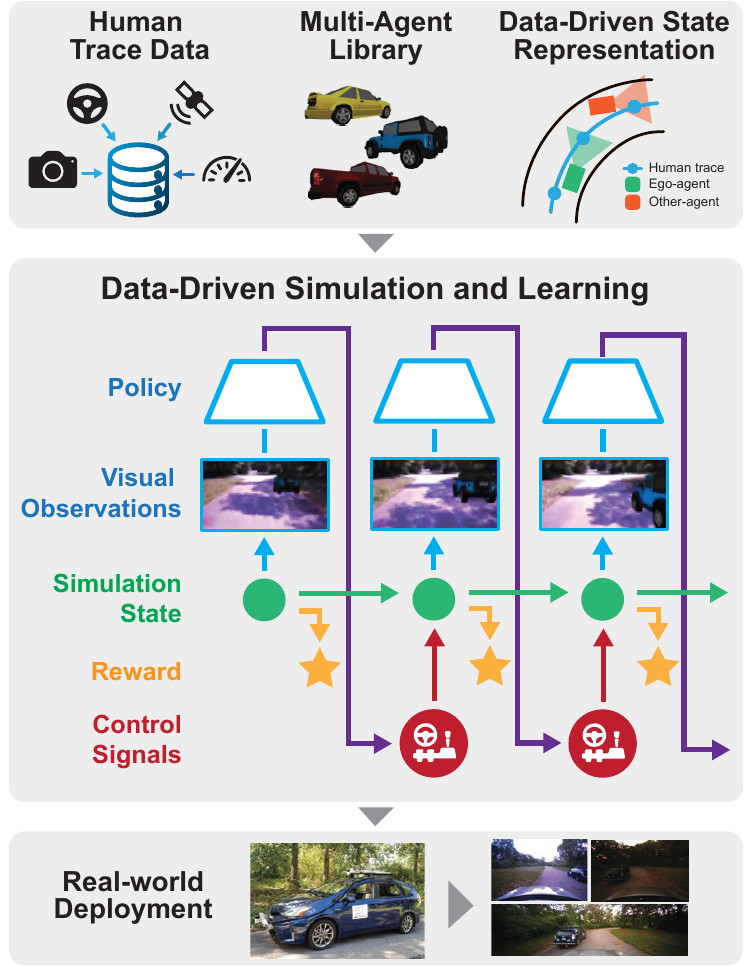}
\caption{\textbf{Data-driven multi-agent simulation.} End-to-end training and testing of control policies with direct transferability to real-world robotic platforms.}
\label{fig:teaser}
\vspace{-15pt}
\end{figure}

In this paper, we present an end-to-end framework for photorealistic simulation and training of autonomous agents in the presence of both static and dynamic agent interactions. Our training environment is photorealistic and supports high-fidelity rendering of multiple agents such that ego-agent learned control policies can be directly transferred onboard a full-scale autonomous vehicle in the real world, without requiring any degree of domain randomization, augmentation, or fine-tuning. Our simulator is fully data-driven: from a single real-world human driving trajectory, virtual agents can synthesize a continuum of new viewpoints from their simulated position. In addition to perceiving the synthesized environment, the simulation system supports the inclusion and rendering of other agents along with their motion dynamics. Several tasks of increased multi-agent interaction complexity are modeled in simulation and used to train policies that can later be deployed directly on physical autonomy platforms such as autonomous cars.  


By simulating arbitrary agent interactions, we do not require massive amounts of expert training data and dense supervisory signals, which are two common limitations of existing imitation learning approaches. Furthermore, by formulating training as a reinforcement learning (RL) problem, we allow the ego agent to autonomously explore the virtual environment and learn to recover from out-of-distribution edge cases and near collisions. We extensively evaluate our method through closed-loop deployment on challenging real-world environments and agents not previously encountered during training. Beyond exhibiting direct transferability, our models demonstrate high performance and generalizability on complex tasks such as autonomous overtaking and avoidance of a partially observed dynamic agent.

In summary, the key contributions of this paper are:
\begin{enumerate}
    \item A method for jointly synthesizing novel multi-agent viewpoints of an environment enabling interaction modelling and simulation.
    \item Adaptation of state-of-the-art policy learning techniques to learn robust control policies for interactive driving tasks such as vehicle following and overtaking.
    \item Real-world experiments with zero-shot policy transfer on a full-scale autonomous vehicle. This includes an extensive evaluation demonstrating successful overtaking behavior on a perceptually challenging test track not seen during training.
\end{enumerate}


\section{Related Work}

%
%
{\bf Model-Based Simulators.}
Most simulation engines for reinforcement learning and robotics focus on modelling the underlying physics~\cite{Coumans2021,Tassa2020,Todorov2012}.
Recently, simulators based on video-game engines (such as  \textit{CARLA}~\cite{Dosovitskiy2017},  \textit{AirSim}~\cite{Shah2017}, or \textit{FlightGoggles}~\cite{Guerra2019}) also offer a high-fidelity stream of visual data mainly focusing on navigation tasks. 
However, synthetically generated images are typically still insufficient to enable zero shot transfer of learned policies and require techniques such as domain-randomization~\cite{Loquercio2020,tobin2017domain} or style transfer~\cite{Pan2017} to enable real-world deployment of learned policies.
In contrast, the present work follows the philosophy of data-driven simulation to enable zero-shot policy transfer to real-world platforms.

%
%
{\bf Data-driven Simulation.}
A philosophically different approach for achieving the goal of visually and physically realistic simulators is using real-world data for building the simulator.
This is inspired by elaborate data augmentation~\cite{AbuAlhaija2018}, weakly-supervised learning~\cite{Remez2018}, and generative video manipulation~\cite{Menapace2021} techniques. Typically this sacrifices some of the environment editing abilities in favor of maintaining photo-realism.
Improving learning for autonomous driving~\cite{Li2019} is one of the main applications of these ideas focusing on individual sensors such as camera~\cite{Chen2021} or LiDAR~\cite{Manivasagam2020,Wang2020a}.
Simulators such as \textit{Gibson}~\cite{Xia2018}, \textit{Habitat}~\cite{Savva2019}, or \textit{RoboTHOR}~\cite{Deitke2020} offer reconstructed real-world environments for embodied AI research.
In a similar spirit, the main focus of the present work is enabling zero-shot transfer for learning multi-agent interactions.

%
%
{\bf Navigation Policy Learning.}
Traditional planning and control approaches for navigation~\cite{Alcala2020,Carrau2016,Galceran2017,Liniger2019,Schwarting2018} typically do not make use of high dimensional inputs and use  game-theoretic approaches for interaction modelling~\cite{Schwarting2019,Williams2017,Liniger2020, Schwarting2019a}. 
End-to-end navigation approaches can learn driving policies from image inputs~\cite{Chen2020,Liang2018}.
This, however, usually requires human supervision~\cite{Pomerleau1989, Bojarski2016, Pfeiffer2017, Amini2018, Toromanoff2018, Codevilla2018, Amini2019, Bansal2019, Lechner2020, Hawke2020, Zhu2020}, does not consider interactions~\cite{Amini2020,Bewley2019,Brunnbauer2021}, or does not consider deployment on real-world platforms~\cite{Sauer2018,Amini2018a,Song2021,Schwarting2020,Toromanoff2020}.
The present work addresses these limitations demonstrating interactive driving with policies purely trained in simulation.


\section{Methodology}
\begin{figure*}[t!]
    \centering
    \includegraphics[width=\linewidth]{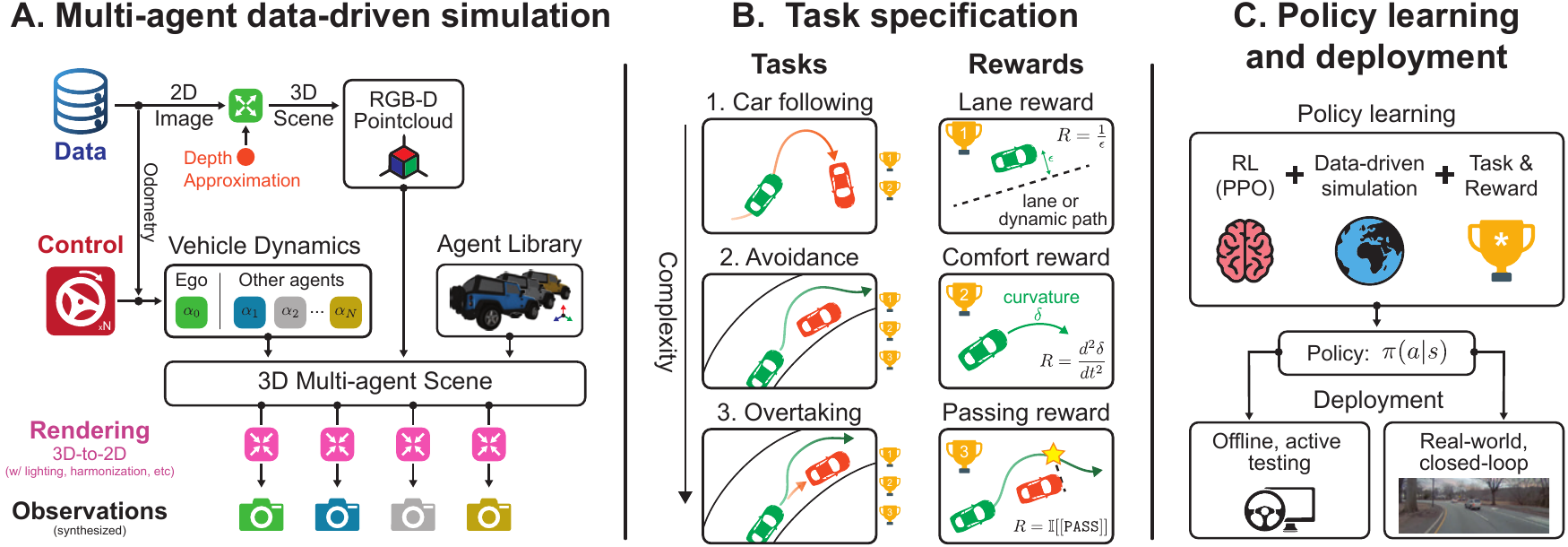}
    \caption{\textbf{Learning and deployment pipeline}. (A) Agent dynamics are simulated within a virtual 3D  environment world, which is generated from real 2D-image data. Photorealistic observations are rendered from the point-of-view of each agent. (B) Tasks with increasing levels of multi-agent interaction are defined within the virtual simulation world. (C) A control policy for a given task is optimized within the simulator using RL and is evaluated both offline as well as online on a full-scale autonomous vehicle in the real-world. }
    \label{fig:overview}
    \vspace{-10pt}
\end{figure*}
In the following section, we outline our simulation and training pipleine (\figref{fig:overview}) starting with the creation of a multi-agent simulator and photorealistic synthesis of novel perception viewpoints. We then discuss how such an environment can serve as a flexible backbone for optimizing RL policies by defining control tasks which encourage multi-agent interaction. 
\subsection{Data-driven Simulation and Learning Environment}
\begin{figure}[b!]
    \centering
    \vspace{-10pt}
    \includegraphics[width=\linewidth]{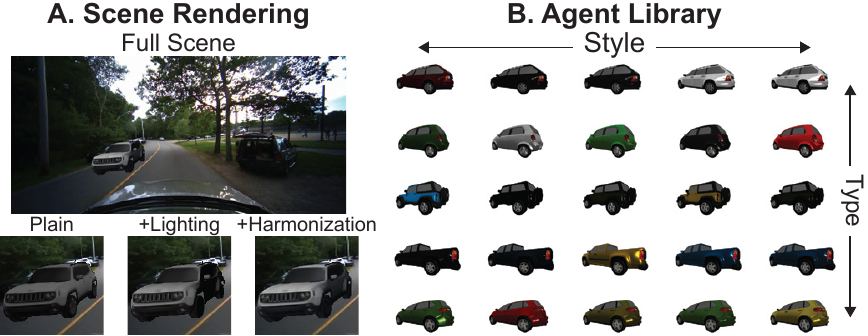}
    \caption{\textbf{Multi-agent rendering.} (A) Example simulated scene with an oncoming agent. Different levels of rendering compare the effect of lighting and harmonization. (B) A variety of agents are available during simulation with different styles, body material, specularity, and color.}
    \label{fig:rendering}
\end{figure}

The high-level pipeline of the proposed multi-agent data-driven simulation consists of (1) updating states for all agents, (2) recreating the world by projecting real-world image data to 3D space based on depth information, (3) configuring and placing meshes for all agents in the scene, (4) rendering the agent's viewpoint, and (5) post-processing in image space to add lighting and to harmonize the image. The vehicle dynamics follow the continuous kinematic model
\begin{align}
    [
        \dot{x}, \dot{y}, \dot{\phi}, \dot{\delta}, \dot{v}
    ]^\top = \left[
        v\cos{\phi}, v\sin{\phi}, \frac{v}{L}\tan{\delta}, u^{\delta},  u^{a}\right]^\top,
\end{align}
where $x,y, \phi$ are vehicle position and heading. $\delta, v$ are the steering angle and speed, $L$ is the inter-axle distance, and $u^{\delta}, u^{a}$ are the steering velocity and acceleration. We ensure accuracy of discrete-time dynamics equivalent by integration through a third-order Runge-Kutta scheme. We follow \cite{Amini2020} to recreate the virtual world from real data. The pre-collected image sequences represent sparsely-sampled trajectories of perception observations from the real world, which allow photorealistic and semantically accurate local view synthesis around the viewpoint of each frame. Every agent's current state is associated with a pre-collected frame according to the closest distance from its pose to the pose of the data-collection vehicle. We compensate for the local transform between the data and the agent leveraging the camera extrinsics during rendering and using approximate depth information. We project the image from the data coordinate frame, i.e. the data viewpoint $p_s$, to the agent's viewpoint $p_t$,
\begin{align}
    p_t = K T_{v_2\rightarrow t}T_{v_1\rightarrow v_2}T_{s\rightarrow v_1} D_s(p_s)K^{-1} p_s,~~T_{s\rightarrow v_1}=T_{v_2\rightarrow t}^{-1},
    \label{eq:cam_proj}
\end{align}
where $K$ is the camera intrinsic, $D_s(p_s)$ is the depth of the data at point $p_s$, $T_{s\rightarrow v_{*}}$ is the transform from camera to vehicle body, and $T_{v_1\rightarrow v_2}$ is the transform from the vehicle body of the data to that of agent. We explicitly model the transform using vehicle body $T_{v_1\rightarrow v_2}$ since it simplifies placing new meshes in the scene and checking for collisions among meshes. Note that the yaw difference in $T_{s\rightarrow v_{*}}$ may induce a bias of how the mesh is placed towards the left or right, which is of great importance with other agents' present in the scene. Each agent is embodied by a mesh randomly sampled from a parametrizable vehicle mesh library, which allows randomization over different car models, a set of physically-realistic diffuse color, specular color, specular highlight, metalness and roughness of car body material, as shown in \figref{fig:rendering}B. The mesh is reconfigured at the beginning and remains the same throughout an entire episode. We place the mesh based on the relative transform between the corresponding agent and the egocentric viewpoint for rendering. For lighting, we cast ambient light based on the average color of the scene and directional light from the egocentric viewpoint infinitely far away. For postprocessing in image space, we run image harmonization \cite{Sofiiuk2021} with the rendered images and the foreground mask obtained via the rendering process.
%
%
\subsection{Multi-agent Objectives and Tasks}
\noindent\textbf{Tasks for autonomous driving.} We showcase end-to-end policy learning with the proposed multi-agent data-driven simulation on two autonomous driving tasks, \textit{car following} and \textit{overtaking}, majorly focusing on the latter one as it is more challenging. Both tasks involve two agents in the scene, the ego car and the front car. The front car is randomly initialized at some distance along the forward direction of the ego car with random lateral shift and heading with respect to the road curvature. The objective of car following is to keep track of the front car while the front car may perform lane changes. The ego car and the front car are set to a random speed but the same for both so that the ego car never loses the front car in its viewpoint and has sufficient information to perform tracking. In overtaking, the speed of ego car is set to a larger value than that of the front car, allowing the ego car to catch up and overtake. The goal of the ego car is to pass the front car without collision while performing lane stable maneuver. The front car of both tasks use pure pursuit controller to trace out trajectories automatically generated based on road curvature, while the ego car commands steering.

\noindent\textbf{Environment for reinforcement learning.} We formulate our end-to-end policy learning as a RL problem given its generalizability across a wide variety of tasks. Setting up a RL environment requires definitions of environment dynamics, terminal condition, and reward function. The simulator defines how the scene dynamically evolves after receiving an agent's action. Due to the nature of the simulator rendering, a default set of terminal conditions is exceeding a threshold maximum translation or rotation, as discussed in \cite{Amini2020}. Another terminal condition occurs when there is collision among agents, determined by the overlap of polygons with shape as vehicle dimension exceeding certain threshold. Besides, given how new observation is synthesized based on a local transform with respect to the data in the simulator and all autonomous driving tasks involving lane following, it's natural to define reward function based on rotational, lateral, and longitudinal component of vehicle pose with respect to the data (center line).
\begin{align*}
    \begin{bmatrix}
        q_{lat} \\ q_{long} \\ q_{rot}
    \end{bmatrix} = \begin{bmatrix}
        \cos{\theta_s} & -\sin{\theta_s} & 0\\
        \sin{\theta_s} & \cos{\theta_s} & 0 \\
        0 & 0 & 1
    \end{bmatrix}\begin{bmatrix}
        x_t - x_s \\ y_t - y_s \\ \theta_t - \theta_s
    \end{bmatrix},
\end{align*}
where $(x_t, y_t, \theta_t)$ and $(x_s, y_s, \theta_s)$ are the poses of the virtual agent and corresponding human reference respectively. With this notation, the two aforementioned terminal conditions can be written as $\mathbbm{1}_{q_{lat}>Z_{lat}}$ and $\mathbbm{1}_{q_{rot}>Z_{rot}}$, where $Z_{*}$ is the threshold that triggers termination. We define a lane following reward as
\begin{align*}
    R_{lane} = 1 - \left(\frac{q_{lat}}{Z_{lat}}\right)^2.
\end{align*}
For car following, we can simply adapt the lane reward by changing the center line to the trajectory traced out by the front car. In overtaking, additional to lane reward, we define a pass reward based on comparing the distances traced out by both cars,
\begin{align*}
    R_{pass} = \mathbbm{1}\left[\int \sqrt{\dot{x}_{e}(t)+\dot{y}_{e}(t)} - \int\sqrt{\dot{x}_{f}(t)+\dot{y}_{f}(t)}\geq Z_{pass}\right],
\end{align*}
where subscripts $*_e$ and $*_f$ denote ego and front car respectively, and $Z_{pass}$ is hyperparameter. To provide more learning signal for collision avoidance, we dilate the polygon of the ego car and compute its overlap with other agents,
\begin{align*}
    R_{collision} = -\frac{|\text{Dilate}(P_{ego})\cap P_{other}|}{|P_{ego}|},
\end{align*}
where $P$ denotes a vehicle's polygons. In both tasks, we add a comfort reward that is computed as the negative second derivative of steering $R_{comfort}=-\ddot{\delta}$ to reduce jittering.

\noindent\textbf{Policy learning.} While there is no limitation for RL algorithms to be applied in our simulator, we adopt proximal policy optimization (PPO) \cite{Schulman2017} for its ubiquity and simplicity. PPO is an on-policy algorithm that maximizes the objective,
\begin{multline*}
    \underset{s,a\sim\pi_k}{\mathbbm{E}}\Big[\min(\frac{\pi_{k-1}(a|s)}{\pi_{k}(a|s)}A^{\pi_k}(a|s),\\
    \text{clip}(\frac{\pi_{k-1}(a|s)}{\pi_{k}(a|s)}, 1-\epsilon,1+\epsilon)A^{\pi_k}(a|s))\Big],
\end{multline*}
where $\pi(a|s)$ is the policy's action distribution given observation $s$, $A^{\pi_k}(a|s)$ is advantage function which estimates how good an action is, and $\epsilon$ is a hyperparameter. The intuition is to maximize task performance, measured by the advantage function, while making sure the new policy $\pi_k$ does not deviate too much from the old policy $\pi_{k-1}$ by bounding the ratio $\pi_{k-1}/\pi_k$ to a small interval, $\epsilon$. In our case, the policy takes observation $s$ as images and outputs action $a$ as steering angle. We adopt a convolutional neural network (CNN) to extract image features, followed by a long short-term memory (LSTM) recurrent network to capture motion information of other agents in the scene.



\section{Results}
\label{sec:result}

\subsection{Data Collection and Experimental Setup}

\noindent\textbf{Real-world testbed.} We deploy the trained policies onboard our full-scale autonomous vehicles (2019 Lexus RX 450H) which we equipped with a NVIDIA 2080Ti GPU and an AMD Ryzen 7 3800X 8-Core Processor. The primary perception sensor is a 30Hz BFS-PGE-23S3C-CS camera, with a $130^{\circ}$ horizontal field-of-view. Image data is serialized with a resolution of $960 \times 600$. Other onboard sensors include inertial measurement units (IMUs) and wheel encoders, which provide curvature feedback and odometry estimates, as well as a global positioning system (GPS) for evaluation purposes. 
Data for simulation is collected at a speed of $\sim 30$ kph while for autonomous evaluation, the vehicle speed is lowered to $\sim 15$ kph for safety reasons.

\noindent\textbf{Implementation details for policy learning.}
At episode initialization, the front car is placed randomly $6-15$m in front and $1-2$m laterally shifted relative to the ego car. During simulation, the front car uses a pure pursuit controller from the reference path with gain 0.8 and lookahead distance as 5m. The overlap threshold for determining collision is 5$\%$ during training and 0$\%$ at evaluation. We dilate the ego car by 1m in length and 0.4m in width and adopt episode level data augmentation by perturbing image gamma, brightness, saturation, and contrast. Images are standardized before fed to the CNN. We follow default parameters for PPO \cite{Schulman2017}, except for a larger buffer size 32000 and minibatch size 512 to increase training stability.

\noindent\textbf{Evaluation metric.} We follow two types of evaluation protocols: \textit{online active test} and \textit{offline active test}. Online test runs end-to-end control policies on physical full-scale autonomous vehicle, while offline test runs policies in the simulator. These tests examine the transferability from our photorealistic simulator to real-world. Active (i.e. closed-loop) testing allows policies to take action and interact with the environment as opposed to passive test that computes difference from control command pre-collected in demonstration. Active testing presents a significantly more challenging evaluation and is preferred here due to the inability of passive, open-loop testing to effectively evaluate a policy~\cite{Codevilla2018Passive}. In complicated scenarios and interactions with multiple agents in a scene, it is ambiguous to define ground truth for maneuvers, e.g., the policy can perform overtaking at any time as long as it keeps sufficient clearance from the front car, resulting in an infinite numbers of trajectories qualified to be considered as ground truth. 

We measure the performance of our end-to-end policies from two perspectives: \textit{safety} and \textit{stability}. For a safety metric, we compute the \textit{rate of intervention} while autonomous. In offline test, intervention is determined by whether the overlap of agents is larger than zero. In online test, intervention is determined by whether the human test driver takes over control of the vehicle. An additional safety metric is \textit{minimal clearance}, which estimates the shortest distance between polygons that represent vehicle throughout a trial. Stability metrics measure the ability of the policy to perform lane stable maneuver during overtaking. We compute the \textit{maximal deviation}, the largest lateral shift from the lane center, and \textit{maximal yaw}, the largest yaw difference from road curvature, throughout a trial. The ideal maneuver finds a perfect trade-off between these two metrics types. 

\subsection{Policy Learning and Offline Testing}
%

%
%
In this section, we present policy learning in the proposed multi-agent data-driven simulator and extensive analysis from offline active test. 
Compared to existing imitation learning approaches, which require vast amounts of data to generalize to real-world driving~\cite{Bojarski2016, Amini2019, Hawke2020}, our system demands significantly less data due to its ability to synthesize a continum of different viewpoints and trajectories from a single driving trace. Specifically, we collect around 30 minutes data in an indoor garage and around 1 hour in rural roads under different weather and time of day as dataset used for our simulator.
The garage data contains only straight lanes under fixed lighting. Data from outdoor rural roads contains a wide variety of road curvatures and lighting conditions with glares and shadows, presenting significant challenges for image-based controllers. 
\begin{wrapfigure}[12]{r}{0.25\textwidth}
\centering
\includegraphics[width=0.25\textwidth]{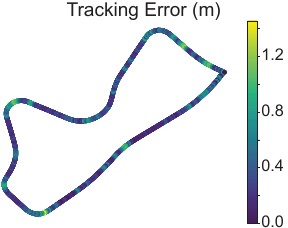}
\caption{Car following tests.}
\label{fig:car_following}
\end{wrapfigure}
In \figref{fig:car_following}, we evaluate our car-following policy through 100 simulated episodes on rural road data, each randomly initialized at a random position and tracing out a segment in the test track, and compute average tracking error at every position of the track. While the tracking error increases generally due to random lane change of the front car, we can still observe greater deviation at positions with large road curvature. 
%
\begin{table}[b!]
\centering
\resizebox{1\linewidth}{!}{
    \begin{tabular}{@{}c|cccc@{}}
    \toprule
    \textbf{Environment} &
      \begin{tabular}[c]{@{}c@{}}\textbf{Average}\\ \textbf{Intervention $\shortdownarrow$}\end{tabular} &
      \begin{tabular}[c]{@{}c@{}}\textbf{Minimum}\\ \textbf{Clearance $\shortuparrow$}\end{tabular} &
      \begin{tabular}[c]{@{}c@{}}\textbf{Maximum}\\ \textbf{Deviation $\shortdownarrow$}\end{tabular} &
      \begin{tabular}[c]{@{}c@{}}\textbf{Maximum}\\ \textbf{Yaw $\shortdownarrow$}\end{tabular} \\ \midrule
    Garage &
      0.129 &
      0.216 &
      0.361 &
      0.112 \\
    Outdoor &
      0.167 &
      0.240 &
      0.360 &
      0.082 \\ \bottomrule
    \end{tabular}
}
\caption{Offline overtaking policy evaluation.}
\label{tab:offline_active}
\end{table}
To evaluate offline overtaking policies, we run 1000 simulated episodes in both the garage and rural environments (\tabref{tab:offline_active}). The rate of intervention in the garage is lower since it is an easier environment overall due to consistent lighting. The minimal clearance throughout an episode is on average around 20cm. The maximal deviation and yaw are also both relatively low to allow for overtaking without exceeding maximal lateral shift $Z_{lat}$, which deteriorates the local view synthesis. 

\begin{figure}
    \centering
    \includegraphics[width=1\linewidth]{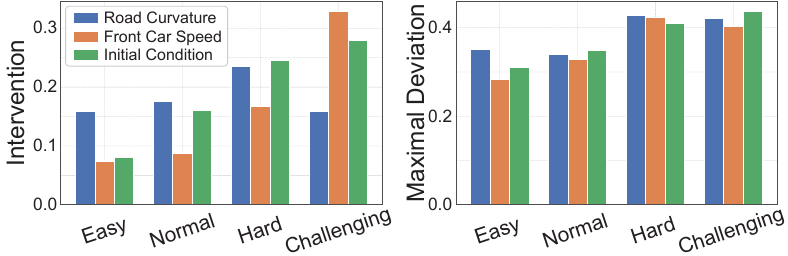}
    \caption{\textbf{Error analysis for overtaking.} Breakdown of how the rate of interventions (left) and maximal deviation (right) change as a function of the trial complexity. }
    \label{fig:diff_config}
\end{figure}

\begin{figure}
    \vspace{-10pt}
    \centering
    \includegraphics[width= 1.0\linewidth]{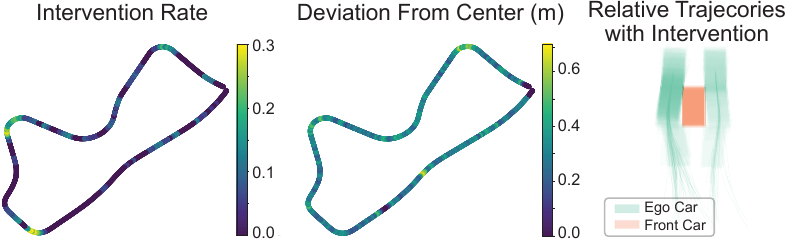}
    \caption{\textbf{Overtaking performance analysis.} Rural intervention rates (left), and deviation from center (middle). Visualization of trajectories from intervention episodes (right).}
    \label{fig:crash_devens_loop}
\end{figure}

In \figref{fig:diff_config}, we perform error analysis on different configurations, including road curvature, front car speed, and initial condition (lateral difference between two cars). We group each configuration according to 4 levels: easy, normal, hard, and challenging. We can see that higher car speed and challenging initial condition greatly increase intervention rate, while the effect of larger road curvature is inconclusive, potentially due to unbalanced number of samples. The results of maximal deviation is consistent with intervention. To further investigate the effect of road curvature, in the left of \figref{fig:crash_devens_loop} we run 1000 episodes, with every position of an segment tagged by whether the corresponding episode involves intervention, allowing each position at the track have its own normalized measure of intervention rate. We can see increased number of intervention at region with high road curvature. In the middle of \figref{fig:crash_devens_loop}, the average deviation from the center spreads out almost evenly across the test track. In the right of \figref{fig:crash_devens_loop}, we overlay the relative trajectories with respect to the front car and the polygons of both vehicles at the last step for episodes with intervention. Most interventions result from slight sideswipes, implying the end-to-end policy still perform reasonable maneuvers even in failure cases. 

\begin{figure}
    \centering
    \includegraphics[width= 1.0\linewidth]{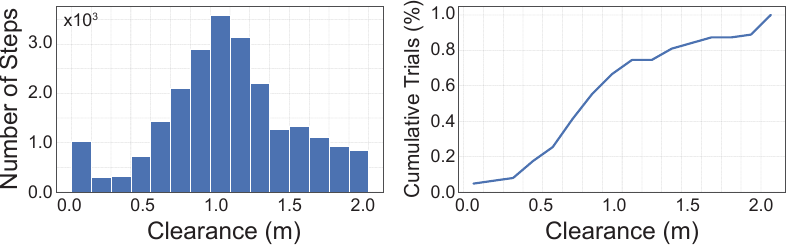}
    \caption{\textbf{Clearance analysis in real-world tests.} Histogram of clearance across \textit{steps} (left), and normalized cumulative \textit{trials} recall at clearance (right). }
    \label{fig:clearance_analysis}
\end{figure}

\subsection{Real-world Testing and Evaluation}

\begin{table}[b!]
\centering
\resizebox{1\linewidth}{!}{
    \begin{tabular}{@{}cc|cccc@{}}
    \toprule
    \multicolumn{2}{c}{\textbf{Scenario}} &
      \begin{tabular}[c]{@{}c@{}}\textbf{Average}\\ \textbf{Intervention $\shortdownarrow$}\end{tabular} &
      \begin{tabular}[c]{@{}c@{}}\textbf{Minimum}\\ \textbf{Clearance $\shortuparrow$}\end{tabular} &
      \begin{tabular}[c]{@{}c@{}}\textbf{Maximum}\\ \textbf{Deviation $\shortdownarrow$}\end{tabular} &
      \begin{tabular}[c]{@{}c@{}}\textbf{Maximum}\\ \textbf{Yaw $\shortdownarrow$}\end{tabular} \\ \midrule
    \multirow{2}{*}{\begin{tabular}[c]{@{}c@{}}Front\\ Car\end{tabular}}     & Static  & 0 / 33 (0.00) & 1.092 & 1.341 & 0.783 \\
                                                                             & Dynamic & 3 / 30 (0.10) & 1.063 & 1.453 & 0.332 \\ \midrule
    \multirow{2}{*}{\begin{tabular}[c]{@{}c@{}}Overtake\\ From\end{tabular}} & Left    & 0 / 32 (0.00) & 1.410 & 1.961 & 0.573 \\
                                                                             & Right   & 3 / 31 (0.09) & 0.735 & 0.809 & 0.563 \\ \bottomrule
    \end{tabular}
}
\caption{Online (real-car) active test.}
\label{tab:online_active}
\end{table}

\begin{figure*}
    \centering
    \includegraphics[width=0.95\linewidth]{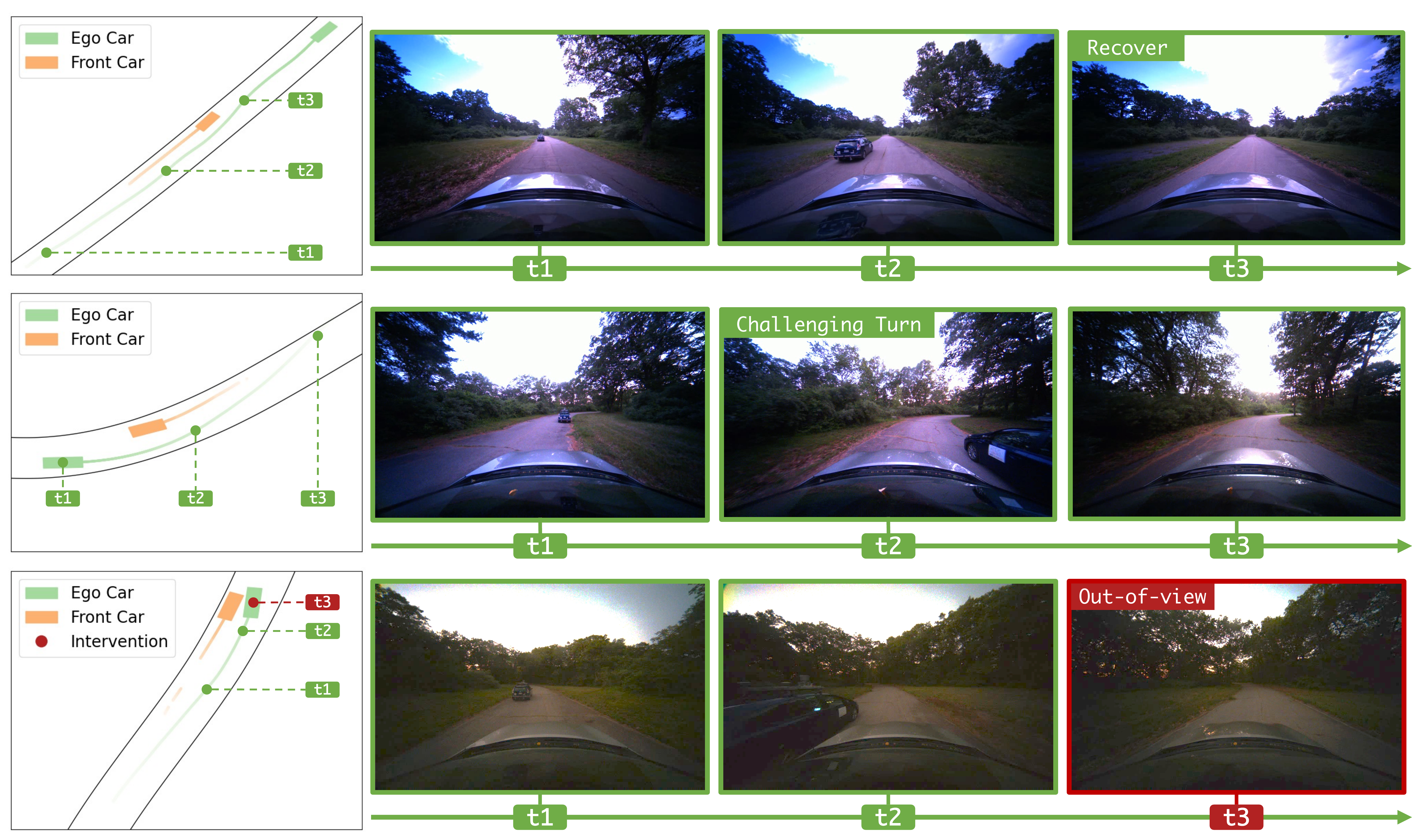}
    \caption{\textbf{Qualitative real-world inspection.} Rows correspond to individual trials from different locations on on the test track. GPS tracking data is visualized (left) along with timepoints tagged. For each timepoint, the corresponding image view is provided (right) along with a semantic description.}
    \label{fig:qual_results}
\end{figure*}

To further verify the effectiveness of the end-to-end policy learned from our simulator, we deploy the learned model onto our full-scale autonomous vehicle and carry out in total 63 trials (2 hours autonomously) for obstacle avoidance and overtaking. In \tabref{tab:online_active}, we show overall performance for difference scenarios. Front car being dynamic poses a more challenging task in comparison to being static and results in more intervention. In comparison to offline test, minimal clearance, maximal deviation and yaw are larger since the agent was placed in a more challenging setting by initializing at either extreme side of the road and starting the ego car immediately behind it. 
We also find the intervention frequency and minimal clearance to be biased towards overtaking from the right side. Our hypothesis is that the yaw difference between camera and vehicle slightly drifts along time between the real car test and camera calibration for the simulator. Even 1 degree drift can cause $\sim$9cm lateral shift for objects 5m in the front of the camera.

We perform clearance analysis in \figref{fig:clearance_analysis}. We show the histogram of steps with clearance less than 2m across all episodes. The peak is at 1m, which is good for safe maneuver. Observing the local maximum at very small clearance, we evaluate the normalized cumulative trials that recall different clearance thresholds. The small initial slope indicates that most low clearance steps are contributed by a very small proportion of trials. Besides, the largest slope occurs roughly close to 0.8m, indicating that most trials have clearance at such distance. In \figref{fig:qual_results}, we demonstrate qualitative results with two of them performing successful overtaking and one involving intervention. The first example (row) shows how the end-to-end learned policy can avoid the moving front car and recover to the lane. In the second row, we further showcase the capability of our policy to make a challenging turn, where the road curves toward the right while the ego car can only overtake from the left. In the last row, we conduct a case study involving intervention. At the first two timesteps $t1$ and $t2$, the policy did manage to avoid the front car, while at $t3$, it cut back too early since the front car is lost in the view and the ego car speed is too slow for the recurrent model to memorize there was a car some time ago in the left. A potential straightforward solution can be augmenting more cameras on the side of the autonomous vehicles.



\section{Conclusion}
\label{sec:conclusion}
In this paper, we present a novel method to learn an end-to-end controller using multi-agent data-driven simulation for autonomous driving. We propose several multi-agent tasks with increasing levels of complexity and conduct extensive empirical analysis within simulation as well as the real-world, where our learned policy is deployed onboard a full-scale autonomous vehicle. By leveraging photorealistic simulation we drastically reduce the amount of data required by our agent to learn a transferable policy. Potential future directions include but are not limited to multi-modality simulation, more complicated tasks that involve situational awareness from agent interaction, and edge case generation by evaluation with adversary. We believe our work opens up a new venue for leveraging high fideltiy data-driven simulation to train and evaluate visual controllers in multi-agent scenarios.





\bibliographystyle{IEEEtran}
\bibliography{refs}  

\end{document}